%% file: main.tex
\documentclass{article}
\usepackage{amssymb}

\usepackage[final]{corl_2025} 

\usepackage{amsmath}
\usepackage{multirow}
\usepackage{graphicx}
\usepackage{cleveref}
\usepackage{booktabs}
\usepackage{xcolor}

\title{ActLoc: Learning to Localize on the Move via Active Viewpoint Selection}

\author{
  Jiajie Li\textsuperscript{1,*},
  Boyang Sun\textsuperscript{1,*},
  Luca Di Giammarino\textsuperscript{2},
  Hermann Blum\textsuperscript{1,4},
  Marc Pollefeys\textsuperscript{1,3} \\
  \textsuperscript{1}ETH Zürich \quad
  \textsuperscript{2}Sapienza University of Rome \quad
  \textsuperscript{3}Microsoft \quad
  \textsuperscript{4}University of Bonn \\
  \textsuperscript{*}Equal contribution \\
  \color{blue}{https://boysun045.github.io/ActLoc-Project/}
}

\begin{document}
\maketitle


\begin{figure*}[h]
    \centering
    \includegraphics[width=1.0\linewidth]{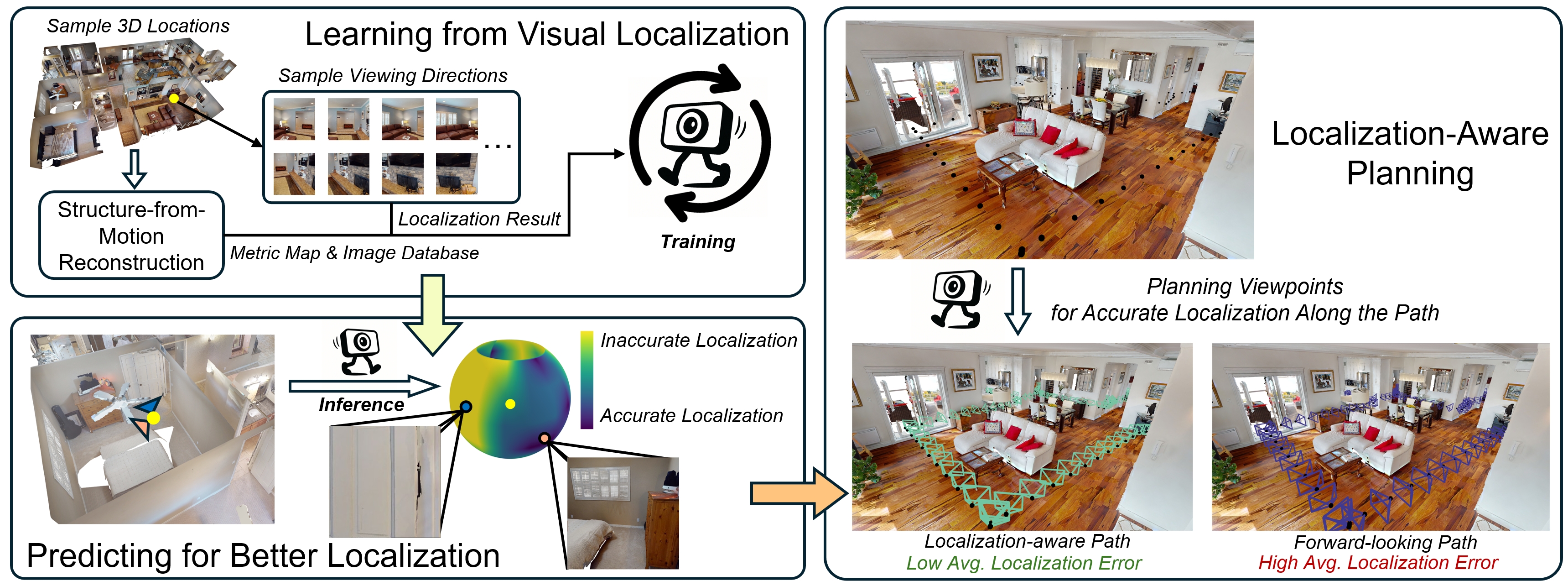}
    \caption{\textbf{ActLoc} is an active viewpoint selection framework designed to improve visual localization for robot navigation. \textbf{Left}: The model is trained using supervision from Visual Localization, learning to predict distributions of localization accuracy over yaw and pitch angles at arbitrary 3D locations. \textbf{Right}: The trained model is integrated into the path planning process, enabling the robot to select viewpoints at each waypoint that increase localization robustness along its trajectory.}

    \label{fig:teaser}
\end{figure*}

\begin{abstract}
  Reliable localization is critical for robot navigation, yet most existing systems implicitly assume that all viewing directions at a location are equally informative. In practice, localization becomes unreliable when the robot observes unmapped, ambiguous, or uninformative regions. To address this, we present \textbf{ActLoc}, an \textbf{act}ive viewpoint-aware planning framework for enhancing \textbf{loc}alization accuracy for general robot navigation tasks. At its core, ActLoc employs a large-scale trained attention-based model for viewpoint selection. The model encodes a metric map and the camera poses used during map construction, and predicts localization accuracy across yaw and pitch directions at arbitrary 3D locations. These per-point accuracy distributions are incorporated into a path planner, enabling the robot to actively select camera orientations that maximize localization robustness while respecting task and motion constraints. ActLoc achieves state-of-the-art results on single-viewpoint selection and generalizes effectively to full-trajectory planning. Its modular design makes it readily applicable to diverse robot navigation and inspection tasks.
\end{abstract}



\input{secs/introduction}

\input{secs/related}

\input{secs/method}

\input{secs/experiment}

\input{secs/conclusion}
\newpage
\input{secs/limitation}


\acknowledgments{We sincerely thank the reviewers for their thoughtful comments and constructive suggestions. Their detailed feedback helped us clarify the presentation, improve the organization, and better highlight the contributions of this work.}


\bibliography{main}  

\clearpage
\appendix

\input{secs/appendix}

\end{document}

%% file: secs/introduction.tex
\section{Introduction}

In recent years, visual localization has gained significant traction in robotics for two main reasons. First, compared to LiDAR, cameras are far more affordable, reducing sensor costs and making robotic systems more accessible. Second, visual localization extends beyond robotics, being widely adopted in handheld and head-mounted devices where LiDAR sensors are often too bulky and heavy. Visual maps can also be shared across devices, enabling multi-agent spatial alignment~\cite{cramariuc2022maplab, chang21icra_kimeramulti, blum2025crocodl} and facilitating human–robot collaboration~\cite{chen20243d, salunkhe2025intuitive}.

Despite these advantages, visual localization remains more challenging than LiDAR-based positioning~\cite{dellenbach2022ct,ferrari2024mad,pan2024tro}. For example, in the Hilti SLAM Challenge 2023~\cite{nair2024hilti}, the vision-only track achieved barely half the scores of LiDAR-based methods. A key limitation stems from the narrow field of view (FoV) of cameras, which prevents omnidirectional sensing readily available with more expensive sensor setups \cite{di2022md, cwian2025mad}. Since visual localization relies heavily on image features, viewpoints covering feature-rich regions tend to yield more reliable results. This observation has motivated active localization strategies~\cite{lim2023fisher, hanlon2023active, di2024learning}, where viewpoint selection is integrated into the robot’s motion planning. By actively adjusting its position and orientation, a robot can track feature-rich regions and improve localization accuracy along its trajectory.

Active vision methods have progressed from analytical modeling~\cite{lim2023fisher, zhang2020fisher} to deep learning–based approaches~\cite{hanlon2023active, di2024learning, bartolomei2021semantic, hu2024active}. In the context of active localization, learning-based methods improve accuracy but remain computationally expensive, since evaluating localization quality at each viewpoint demands substantial resources—an issue for real-time systems where fast decisions are essential. Moreover, these methods are trained on small datasets, limiting their generalization ability. Finally, most approaches assess viewpoints independently, requiring sparse sampling and decoupling viewpoint selection from the path planning process.

To address these challenges, we propose an efficient and scalable approach. Instead of evaluating viewpoints individually, our model encodes scene information and directly predicts a spherical distribution of localization quality at each 3D waypoint. This allows dense estimation across multiple yaw–pitch directions and supports trajectory optimization that jointly accounts for task objectives and localization robustness. The resulting system is both computationally efficient and suitable for continuous real-time navigation.

In summary, our main contributions are:
\begin{itemize}
\item A learning-based model trained on visual localization tasks that predicts a distribution of localization accuracy over multiple yaw–pitch directions in a single forward pass.
\item \textbf{ActLoc}, an active localization planner that integrates the model into a complete navigation system for online viewpoint selection and trajectory planning.
\item Comprehensive experiments across multiple datasets, evaluating both single-viewpoint selection and long-horizon path planning, where ActLoc consistently outperforms baselines and heuristic methods.
\end{itemize}

%% file: secs/related.tex
\section{Related Work}
\subsection{Visual Localization}
Visual localization refers to estimating the 6-DoF camera pose of a query image within a known environment. Existing approaches can be broadly divided into two categories. The first is the two-step approach, which establishes correspondences between the query and a database of reference images, followed by pose estimation through geometric optimization. These correspondences may be based on sparse visual features~\cite{sattler2011fast,sattler2012improving,sattler2016efficient,sattler2017large,dusmanu2019d2,sarlin2019coarse,sarlin2021back,panek2022meshloc} or dense pixel-level matches~\cite{brachmann2018learning,brachmann2019expert,yang2019sanet,cavallari2019real,li2020hierarchical,tang2021learning,dong2022visual}. The second is the direct approach, also known as pose regression, which estimates the camera pose directly from the input image, typically using learning-based models~\cite{sattler2019understanding,ding2019camnet,yan2022crossloc,wang2020atloc,xue2020learning,moreau2022lens,shavit2021learning,shavit2022camera,chen2022dfnet}. While both approaches have achieved strong results, they remain inherently \textit{passive}: the camera pose is inferred solely from the captured view, without reasoning about which viewpoint would yield the most reliable localization. This limitation has motivated research into \textit{active visual localization}.

\subsection{Active Viewpoint Selection}
Active visual localization extends beyond passive pose estimation by allowing an agent to actively select its viewpoints in order to improve localization reliability. This direction builds upon the broader field of active perception, which has long studied strategies for guiding sensor orientations and robot motion to maximize the informativeness of observations~\cite{roy1999coastal,costante2016perception,zhang2018perception,6631022,fontanelli2009visual,7989531,10161136}. 
By integrating perception with planning, these approaches enable robots not only to gather richer visual cues but also to make decisions that directly impact navigation performance. 

Early studies on active localization focused on designing evaluation metrics to identify informative viewpoints. 
Such metrics are typically geometric, for instance using Fisher information to quantify the visibility and spatial distribution of landmarks~\cite{zhang2020fisher}. 
While conceptually interpretable, these hand-crafted metrics are often environment-specific and 
become computationally expensive when extended to dense viewpoint sampling across large maps. 

To overcome the limitations of hand-crafted criteria, recent work has turned to learning-based formulations. 
One line of work treats pose estimation as a regression problem using convolutional networks~\cite{kendall2015posenet,clark2017vidloc}. 
Another line integrates perceptual models with policy learning to guide localization~\cite{chaplot2018active,fang2022towards,lodel2022look}, 
sometimes augmented with higher-level cues such as semantics~\cite{Bartolomei2021Semantic-awareLearning}. 
While these policy-driven methods enable adaptive viewpoint control, they often rely on reinforcement learning, 
which couples perception tightly with action execution and hinders generalization. A recent alternative formulates viewpoint selection as a classification task~\cite{hanlon2023active}, yielding promising performance but relying on user-annotated supervision and remaining limited to per-location viewpoint decisions without trajectory-level planning. The most relevant recent advance is Learning-Where-to-Look 
 LWL)~\cite{di2024learning}, which represents the state-of-the-art in active localization. LWL trains a lightweight multi-layer perceptron (MLP) to classify viewpoints as suitable or unsuitable for localization, with training labels generated in a self-supervised way from voxel-grid–based sampling and visibility checks. At test time, however, it requires dense viewpoint sampling and extensive preprocessing (including landmark visibility evaluation and feature binning), making inference computationally costly even with GPU acceleration. Although LWL presents qualitative planning results, it does not offer a fully integrated or quantitatively evaluated trajectory-level planning system.

These limitations motivate our approach, which removes voxel-grid constraints, predicts localization performance across multiple viewpoints in a single pass, and integrates these predictions into motion planning to enable active viewpoint selection along a trajectory.

%% file: secs/method.tex
\section{Method}
\label{sec:method}

\begin{figure*}[t!]
    \centering
    \includegraphics[width=0.99\linewidth]{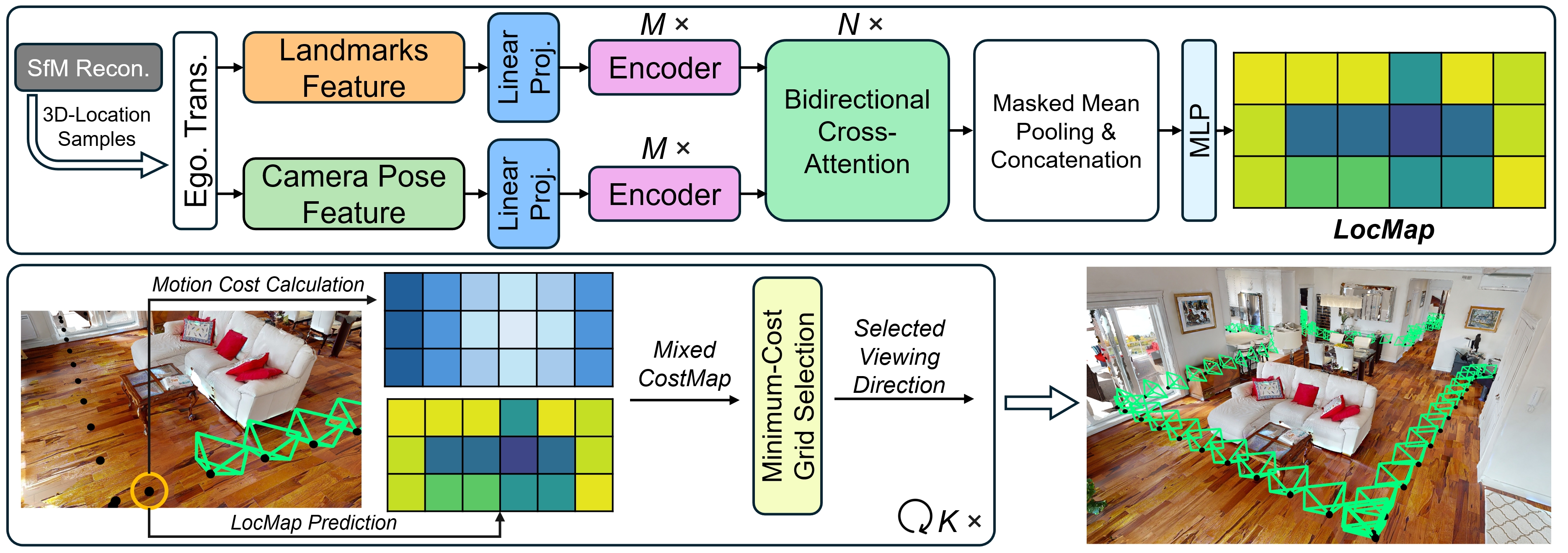}
    \caption{\textbf{System Overview.} \textbf{Top}: The core of ActLoc is an attention-based network that processes multi-modal inputs from an SfM model, including the metric map and camera poses. The network predicts localization performance as a two-dimensional grid over multiple pitch and yaw rotations. \textbf{Bottom}: During path planning, ActLoc constructs a mixed cost map at each candidate location by combining the network prediction with additional factors, such as motion constraints, and selects the viewing direction that minimizes this cost.}

    \label{fig:system}
\end{figure*}

\subsection{System Overview}

Our system consists of two main components: (1) an attention-based model trained on Structure-from-Motion (SfM) data to predict visual localization performance over local yaw and pitch angles. We denote this representation as \textit{LocMap}. (2) The integration of LocMap into a path planning pipeline, which, given a sequence of \textit{waypoints} (robot positions in 3D space), actively selects \textit{viewpoints} (camera orientations at each waypoint) to improve localization accuracy while respecting motion constraints such as continuity. ActLoc predicts the localization outcome for multiple candidate viewpoints in a single forward pass and combines these predictions with planning costs to balance accuracy and feasibility. \Cref{fig:system} provides an overview of the full pipeline.

\subsection{Predicting Localization Performance}
\label{sec:net-output}

We define active localization as the problem of selecting viewpoints along a given sequence of waypoints such that the robot achieves accurate visual localization while following the trajectory. At each waypoint, the planner chooses a viewpoint so that images captured during execution yield reliable localization results. We assume that the environment has been previously mapped, which is standard in visual localization since methods typically query an existing database. This assumption naturally applies to many robotics tasks, such as inspection, monitoring, and indoor service.

To select a viewpoint at a waypoint, the robot must leverage prior knowledge of the scene from mapping and adapt it to the specific location. We therefore train a model that encodes this prior knowledge and predicts the expected localization performance across possible viewing directions. As shown in \Cref{fig:system}, the model outputs the \textit{LocMap}, a two-dimensional grid of size $H \times W$, where each cell represents the predicted localization performance for a candidate viewpoint. The horizontal axis corresponds to yaw rotations, and the vertical axis corresponds to pitch rotations. We exclude extreme pitch angles that point toward the ceiling or floor, as well as in-plane rotations around the camera’s principal axis, since these provide negligible benefit for visual localization~\cite{sp}. Experiments in \Cref{subsec:experiments_on_focal_axis_rotation} further confirm that such rotations often degrade localization performance.

Unlike prior work that either scores only the best viewpoint~\cite{zhang2020fisher} or evaluates multiple viewpoints independently without modeling correlations across orientations~\cite{hanlon2023active,di2024learning}, our model predicts the entire distribution of viewpoint performance in a single forward pass. This design reflects the natural fact that multiple viewpoints may provide good localization, while also avoiding redundant computation from repeated single-viewpoint inference. Moreover, modeling the joint distribution allows the network to capture spatial correlations between viewpoints, which is crucial for accurately predicting localization performance at a waypoint.

We propose an attention-based model to predict the LocMap. This model continuously queries a SfM pipeline to generate training data. Most SfM and visual localization approaches maintain a reconstruction database that includes a metric map and a set of camera poses used for reconstruction. We select these two components as the raw inputs to our pipeline: the 3D landmarks from the metric map $\mathbf{P} \in \mathbb{R}^{3 \times N}$ and the camera poses $\mathbf{T} \in \mathbb{SE}(3)^M$. For a given waypoint $\mathbf{x} \in \mathbb{R}^{3}$, our model predicts the LocMap:

\begin{equation}
    \label{eq:model}
    \mathbf{C}_{\mathbf{x}}^{H \times W} = f_{\theta}(\mathbf{P}, \mathbf{T}, \mathbf{x})
\end{equation}

where $\mathbf{C}_{\mathbf{x}}$ denotes the predicted distribution of localization performance across candidate viewpoints at $\mathbf{x}$. Each cell stores the likelihood that the localization error at the corresponding yaw–pitch direction falls below a threshold. This allows the model to reason about all orientations jointly, instead of evaluating them one by one.


\subsection{Model Architecture}
\label{sec:model}

Our model takes multi-modal input from landmarks and camera poses, encoding each with self-attention layers and fusing them through cross-attention. Given the raw inputs $\mathbf{T}$, $\mathbf{P}$, and $\mathbf{x}$, we apply several preprocessing steps. First, we remove highly uncertain 3D landmarks from $\mathbf{P}$, following~\cite{di2024learning}. Next, we apply an egocentric transformation that shifts $\mathbf{T}$ and the filtered $\mathbf{P}$ from the world frame to the local frame of $\mathbf{x}$. Importantly, we shift only their origins while preserving rotations to maintain consistency with the default viewing direction at $\mathbf{x}$. This implicitly incorporates waypoint location into the inputs without additional modules. Finally, we crop the filtered $\mathbf{P}$ within a bounding box centered at $\mathbf{x}$, enforcing locality and normalizing input scale. The full preprocessing pipeline is shown in \Cref{fig:data_preprocessing_pipeline}.

After preprocessing, camera features are constructed from $\mathbf{T}$ by extracting their Euclidean representation in $\mathbb{R}^7$, consisting of the camera center coordinates and a unit quaternion for orientation $(\mathbf{p_c}, \mathbf{q_c})$. Landmark features are built from the filtered $\mathbf{P}$, with each landmark represented as a 6D vector containing 3D coordinates and normalized RGB values.

Both feature types are independently projected into a shared high-dimensional embedding space using linear layers. Transformer encoder blocks are then applied to each modality, extracting context-aware representations. To integrate modalities, we employ a multi-layer bidirectional cross-attention block, allowing pose embeddings to attend to landmark embeddings and vice versa. This design enables bidirectional information flow and produces fused features that capture the interplay between sparse camera poses and dense 3D landmarks.

Because attention operates on sequences, we apply masked mean pooling to aggregate features into fixed-size vectors while ignoring padding introduced during batching. This yields one vector summarizing pose information and another summarizing landmarks. These are concatenated and passed through an MLP to produce the final output logits. For efficiency, all attention layers use \textit{FlashAttentionV2}~\cite{dao2023flashattention2}, which reduces memory complexity from $O(N^2)$ to $O(N)$ with respect to sequence length and provides substantial speedups, especially for long sequences.

\subsection{Training Data Creation}
\label{sec:training-data}

We train our model entirely on simulated data derived from the HM3D dataset~\cite{ramakrishnan2021habitat}. Specifically, we use the first 100 single-floor scenes, with 90 for training/validation and 10 reserved for testing.

To construct the prior scene maps, we first sample free-space points at typical operation heights between 1.5 and 2.0 meters. At each point, the camera performs a full $360^\circ$ sweep in $36^\circ$ azimuth increments. For each orientation, an integer elevation angle is randomly sampled from $[-15^\circ, 15^\circ)$, and an image is captured. We then run SfM on these images using SuperPoint~\cite{sp} features with the Hierarchical Localization (hloc) framework~\cite{sarlin2019coarse} to obtain 3D landmarks.

To generate training samples, we independently sample waypoints within each scene at heights between $0.4$ and $2.0$ meters, making our approach free of grid-based assumptions as in~\cite{di2024learning}. At each waypoint, the camera rotates horizontally at $20^\circ$ intervals covering the full $[-180^\circ, 180^\circ)$ range. For each azimuth, we additionally vary the elevation from $[-60^\circ, 60^\circ)$ at $20^\circ$ intervals. Finally, we apply hloc and COLMAP~\cite{7780814} to localize each viewpoint against the metric map (3D landmarks), yielding ground-truth localization performance. The full data generation pipeline is illustrated in \Cref{fig:training_data_pipeline}.

\subsection{Path Planning with Localization Awareness}
\label{sec:planning}

LocMap provides a structured representation for viewpoint selection in both single- and multi-viewpoint planning. Unlike prior work that evaluates viewpoints independently, LocMap predicts the distribution of localization performance across all viewpoints at a waypoint in a single inference. This enables optimizing viewpoint choices along a waypoint sequence to achieve higher overall localization accuracy during trajectory execution.

For a given waypoint $\mathbf{x}$, LocMap $\mathbf{C}_{\mathbf{x}}$ provides a direct selection criterion: choosing the grid cell with the highest predicted localization performance. However, during path execution the robot must also satisfy motion constraints, such as limiting rotations between consecutive frames. To address this, we define a mixed cost map that combines localization performance with motion smoothness:  
\begin{equation}
    \mathbf{M}_{\mathbf{x}} = \mathbf{C}_{\mathbf{x}} + \lambda \cdot \mathbf{D}_{\mathbf{x}}
\label{eq:planning}
\end{equation}  
where $\mathbf{D}_{\mathbf{x}}$ is the Mahalanobis distance from each candidate viewpoint to the previously selected viewpoint, and $\lambda$ balances the trade-off between localization accuracy and smooth motion.  

As the robot follows the waypoint sequence, it computes $\mathbf{M}_{\mathbf{x}}$ at each step and selects the viewpoint with the minimum cost. This strategy maintains a balance between accurate localization and motion feasibility, and can be extended to incorporate additional task-specific metrics.

%% file: secs/experiment.tex
\section{Experiments}

We design experiments to address the following questions:  
(a) How well does our model perform in individual viewpoint selection?  
(b) When integrated into ActLoc, how effectively does it balance localization performance with motion constraints?  
(c) How adaptable is ActLoc to different input modality configurations, and how do these configurations perform?  

To evaluate (a), we use test scenes from HM3D and ScanNet~\cite{dai2017scannet} and compare ActLoc against two baselines: FIF~\cite{zhang2020fisher} and LWL~\cite{di2024learning}. \textbf{FIF} is an information-theoretic method based on Fisher Information, which models the camera as a bearing vector to make the measure rotation-invariant. Localization quality is estimated using Gaussian Process regression, and standard information-theoretic metrics, including minimum eigenvalue, determinant, and trace; we report the best-performing metric. \textbf{LWL}, in contrast, employs a lightweight MLP with visibility checks and image binning to generate fixed-length inputs for classifying whether a viewpoint is suitable for localization. We trained LWL on the same dataset for fairness.

To evaluate (b), we integrate the model into the planning pipeline and test it on five HM3D scenes. To better approximate real-world conditions, we sparsify the reconstructed models of these scenes. We compare our approach against LWL and a heuristic forward-facing planner.
All experiments are conducted on a workstation with a single NVIDIA GeForce RTX 4090 GPU.  

\subsection{Single-viewpoint Selection}
\label{sec:single-view}

For the single-viewpoint selection task, we follow the protocol of~\cite{di2024learning}, reporting the percentage of waypoints localized within four thresholds: 0.1m/1°, 0.25m/2°, 0.5m/5°, and 5m/10°. We also report a theoretical upper bound, counting a waypoint as successfully localized if any of its candidate viewpoints meets the threshold.

Our model is configured to output a $6 \times 18$ LocMap, where each cell corresponds to a candidate viewpoint. We adopt this resolution as a trade-off between efficiency and accuracy (see \Cref{subsec:effect_of_LocMap_resolution}). We select the viewpoint with the highest probability of being well-localized. For a fair comparison, LWL is evaluated over the same $6 \times 18$ set of viewpoints, with the best-scoring direction chosen. 

The results in \Cref{tab:comparison_results} show that our model consistently outperforms both baselines at all thresholds. This is particularly noteworthy because our model predicts localization performance across multiple viewpoints simultaneously, which is a more challenging task than the single-viewpoint formulation in LWL, yet it still achieves higher accuracy. These results highlight the model’s ability to capture global spatial structure from SfM inputs and its robustness in generalizing to unseen scenes.

Efficiency is another key advantage. The average processing time per waypoint is 110\,ms for our method, compared to 8230\,ms for LWL (batch mode). This efficiency gap grows further with higher-resolution LocMaps or large-scale scenes, as shown in \Cref{subsec:experiments_on_large_grid_resolutions} and \Cref{subsec:experiments_on_large_scenes}. These results highlight the practicality of our approach for real-time integration into planning pipelines. Additional results in \Cref{sub:robustness_to_sparsification} further show that ActLoc remains robust under various levels of reconstruction sparsity.

\begin{table}[ht]
\centering
\begin{tabular}{@{}llccccc@{}}
\toprule
\toprule
\multicolumn{1}{p{0.01cm}}{} & Method & 0.1m/1° & 0.25m/2° & 0.5m/5° & 5m/10° \\ 
\midrule
\multirow{4}{*}{\rotatebox[origin=c]{90}{HM3D}} 
& LWL \cite{di2024learning} & 90.04 & 90.98 & 91.17 & 91.92 \\
& FIF \cite{zhang2020fisher} & 60.71 & 62.41 & 63.53 & 64.66  \\
& \textbf{ActLoc} & \textbf{92.11} & \textbf{92.48} & \textbf{92.86} & \textbf{93.23} 
\\
& Upper Bound & 97.74 & 97.74 & 97.93 & 98.50 \\
\midrule
\multirow{3}{*}{\rotatebox[origin=c]{90}{Scannet}}
& LWL \cite{di2024learning} & 88.89 & 89.90 & 89.90 & 92.93 \\
& \textbf{ActLoc} & \textbf{91.92} & \textbf{94.95} & \textbf{96.97} & \textbf{96.97} \\
& Upper Bound & 100.00 & 100.00 & 100.00 & 100.00 \\
\bottomrule
\bottomrule
\end{tabular}
\vspace{2pt}
\caption{\textbf{Single-viewpoint Selection Results.} Localization success rates (\%) on ten HM3D and five ScanNet scenes. Metrics follow the Long-Term Visual Localization benchmark~\cite{sattler2018benchmarking}, using three standard thresholds and one additional fine-grained indoor threshold (similar to~\cite{di2024learning}). Our ActLoc model achieves the best results. Best results are highlighted in bold.}

\label{tab:comparison_results}
\end{table}

\subsection{Path Planning Evaluation}

For evaluation in a planning setting, we select several key locations in each scene to ensure coverage of most regions. Path planning is first performed with OMPL~\cite{sucan2012the-open-motion-planning-library} to obtain waypoint sequences, which are then interpolated to generate dense waypoints at 20\,cm intervals. Viewpoint selection is performed using ActLoc, LWL, and a forward-facing baseline. As in the single-viewpoint evaluation, visual localization is executed with hloc using the selected viewpoints. For ActLoc, we tune $\lambda$ in the mixed cost map (Eq.~\ref{eq:planning}), while for LWL we restrict sampling to a local neighborhood around the previously selected viewpoint.  

We evaluate performance using two metrics: success rate and average error across successful localizations. A localization is considered successful if the error is below 5\,m / 10°. Results are summarized in \Cref{tab:greedy_results}. Across all scenes, ActLoc achieves the highest success rate and, in most cases, the lowest average translation and rotation error. Importantly, since LWL requires evaluating every frame, we equalize processing times between ActLoc and LWL for fairness. Under this constraint, LWL is often unable to sample sufficient candidates, resulting in degraded performance.  

\begin{figure*}[h]
    \centering
    \includegraphics[width=0.99\linewidth]{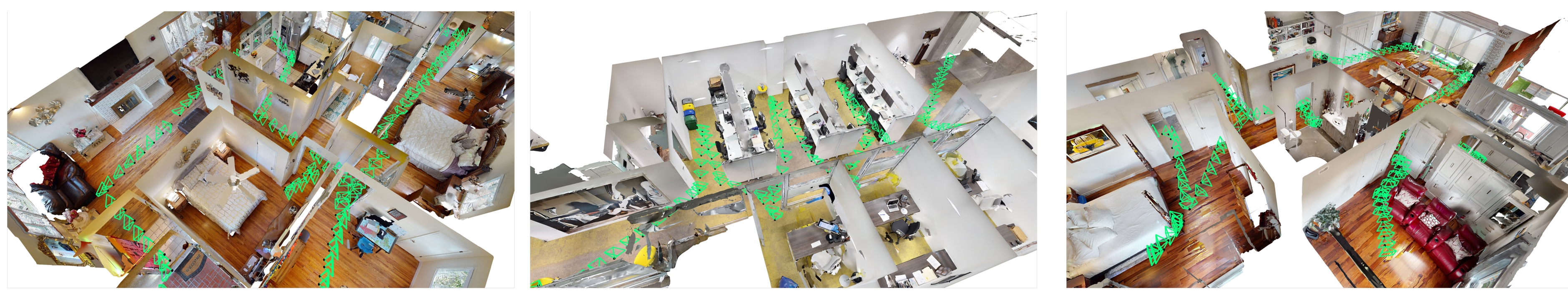}
    \caption{\textbf{Qualitative Planning Results.} Visualization of ActLoc’s viewpoint selection along waypoint sequences across different test scenes. The planned viewpoints demonstrate how ActLoc adapts camera orientation dynamically to balance localization performance and motion smoothness.}
    \label{fig:plan_qual}
\end{figure*}

\Cref{fig:plan_qual} shows qualitative examples of viewpoint selection along waypoint sequences. Thanks to the mixed cost map, ActLoc not only improves localization performance but also maintains smooth rotations between consecutive frames, respecting the robot’s kinematic constraints. These results demonstrate that ActLoc integrates seamlessly into path planning, balancing localization accuracy with motion feasibility while remaining computationally efficient for real-time deployment.  

\begin{table}[ht]
\centering
\begin{tabular}{@{}lccc@{}}
\toprule
Scene ID & Forward & LWL \cite{di2024learning} & ActLoc \\ 
\midrule
00005 & 40.26 / 0.09m / 1.21° & 36.25 / 0.10m / 1.20° & \textbf{52.75} / 0.07m / 1.09° \\
00080 & 11.03 / 0.19m / 1.17° & 3.23 / 0.24m / 1.57° & \textbf{24.00} / 0.05m / 0.80° \\
00105 & 20.10 / 0.22m / 2.34° & 10.54 / 1.13m / 3.78° & \textbf{27.25} / 0.16m / 1.92° \\
00110 & 12.45 / 0.20m / 2.13° & 31.40 / 0.09m / 1.71° & \textbf{35.95} / 0.07m / 1.37° \\
00254 & 29.61 / 0.13m / 1.27° & 27.11 / 0.08m / 1.43° & \textbf{46.08} / 0.14m / 1.62° \\
\bottomrule
\end{tabular}
\vspace{1.4pt}
\caption{\textbf{Quantitative Planning Results.} \textit{Success Rate (\%)/ Translation Error(m) / Rotation Error(°)} across five HM3D test scenes. ActLoc consistently achieves the highest success rate and often the lowest errors, demonstrating its effectiveness in planning settings.}
\label{tab:greedy_results}
\end{table}

\subsection{Ablation Study}

We performed an ablation study to evaluate the individual contributions of camera pose and metric map inputs. Two single-modality variants were tested: ActLoc-Cam, which uses only camera poses, and ActLoc-Map, which uses only point clouds. Both were trained on the same HM3D scenes \cite{ramakrishnan2021habitat} as the full model.

\begin{table}
\centering
\begin{tabular}{@{}lcccc@{}}
\toprule
Method & 0.1m/1° & 0.25m/2° & 0.5m/5° & 5m/10° \\
\midrule
ActLoc-Cam & 84.96 & 86.65 & 87.97 & 88.16 \\
ActLoc-Map & 89.10 & 89.85 & 90.04 & 90.41 \\
\textbf{ActLoc-Full} & \textbf{92.11} & \textbf{92.48} & \textbf{92.86} & \textbf{93.23} \\
\bottomrule
\end{tabular}
\vspace{1.5pt}
\caption{\textbf{Ablation Study.} Localization success rates (\%) at different thresholds for ActLoc-Full, ActLoc-Cam (camera poses only), and ActLoc-Map (point cloud only). ActLoc-Full achieves the best results across all thresholds, confirming the complementary contributions of both modalities.}
\label{tab:ablation_results}
\end{table}

For ActLoc-Map, the camera pose branch and cross-attention blocks were removed, retaining only the landmarks encoder and classifier. ActLoc-Cam used a corresponding simplified architecture focused solely on camera poses.

As shown in \Cref{tab:ablation_results}, ActLoc-Cam achieved the lowest accuracy, while ActLoc-Map performed better but still fell short of the full model. ActLoc-Full outperformed both across all thresholds, confirming the complementary nature of the two modalities and validating the benefit of joint fusion for active viewpoint selection.

%% file: secs/conclusion.tex
\section{Conclusion}
In this work, we introduced ActLoc, an active viewpoint-aware planning framework aimed at enhancing localization accuracy for general robot navigation tasks. ActLoc leverages an attention-based model that predicts localization performance over a dense grid of pitch and yaw angles for arbitrary 3D waypoints, enabling robust viewpoint selection. By integrating these predictions into the path planning process, ActLoc allows robots to dynamically adjust camera orientations to maximize localization robustness while respecting motion and task constraints.

Our experiments demonstrated that ActLoc achieves state-of-the-art performance in single-viewpoint selection and generalizes effectively to full-trajectory planning. Notably, ActLoc provides significant runtime improvements over existing methods, making it well-suited for real-time deployment in practical navigation scenarios.

While our approach shows strong performance, further work is needed to enhance generalization across different feature sets and environments. Future research could explore incorporating data augmentation, normalization strategies, or self-supervised learning to improve robustness. Additionally, extending ActLoc to multi-modal sensor fusion or adapting it for dynamic environments could broaden its applicability to more complex, real-world robotics tasks.

%% file: secs/limitation.tex
\section{Limitations}


The design of ActLoc naturally supports extensions to multi-class discretizations or continuous regression for finer-grained estimation. However, our experiments (\Cref{tab:actloc_multi_results}) show that a naive multi-class formulation with the same training setup does not improve single-viewpoint selection performance. We hypothesize that such extensions may require additional system design and training on larger-scale datasets. At the same time, ActLoc-Multi already provides utility by generating global score heatmaps that reveal spatial variations in localization difficulty (\Cref{sub:global_score_heatmap}). This suggests that finer-grained outputs may still be valuable for higher-level planning tasks, even if they do not directly improve single-viewpoint selection.

We also observe limitations in cross-domain generalization. When using alternative visual features, such as SIFT, ActLoc does not outperform LWL~\cite{di2024learning}, as shown in \Cref{tab:sift_scannet_results}. This highlights a trade-off in our design: while processing raw inputs in a single forward inference enables high efficiency, it may reduce robustness to different feature types. Future work could address this by exploring input normalization or data augmentation strategies.

%% file: secs/appendix.tex
\section{Additional Pipeline Details}
\subsection{Data Generation Pipeline}

\begin{figure}[htbp]
  \centering
  \includegraphics[width=0.99\linewidth]{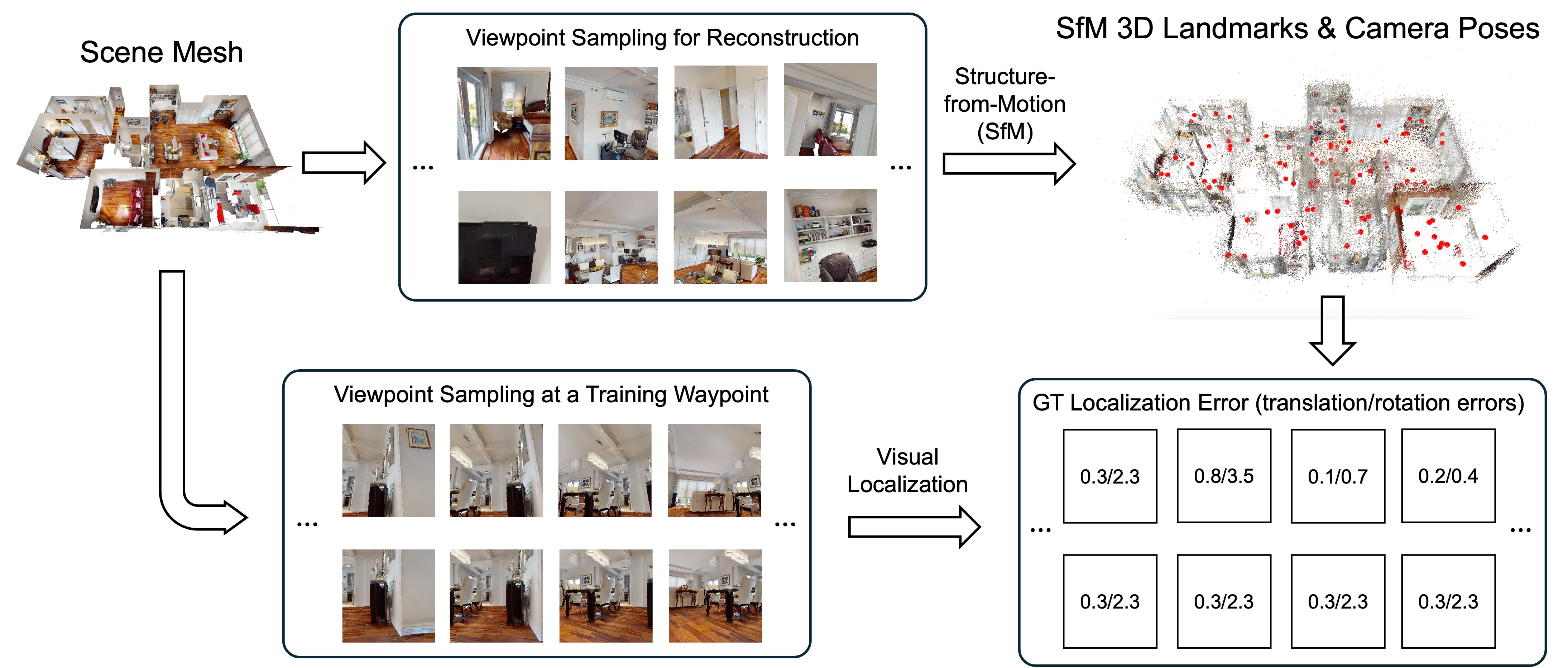}  
  \caption{\textbf{Data Generation.} The pipeline first performs SfM reconstruction on HM3D scenes to obtain 3D landmarks and camera poses. Training waypoints are then sampled throughout each scene, and images captured at multiple yaw–pitch angles are localized against the reconstructed map to produce ground-truth performance labels. This process yields large-scale viewpoint-labeled data for training ActLoc.}
  \label{fig:training_data_pipeline}
\end{figure}

Our training data generation pipeline is shown in \Cref{fig:training_data_pipeline}. 
We use HM3D single-floor scenes as input and first perform SfM reconstruction with SuperPoint features and the hloc framework to obtain a metric map consisting of 3D landmarks and camera poses. 
Next, we sample training waypoints throughout each scene without assuming a grid layout. 
At each waypoint, images are captured by sweeping the camera across multiple yaw and pitch angles. 
Finally, these captured images are localized against the reconstructed metric map using COLMAP, and the resulting errors provide ground-truth labels for training ActLoc. 
This pipeline enables large-scale automatic data generation, ensuring diverse coverage of viewpoints and scene layouts.

\subsection{Data Preprocessing}

\begin{figure}[htbp]
  \centering
  \includegraphics[width=0.99\linewidth]{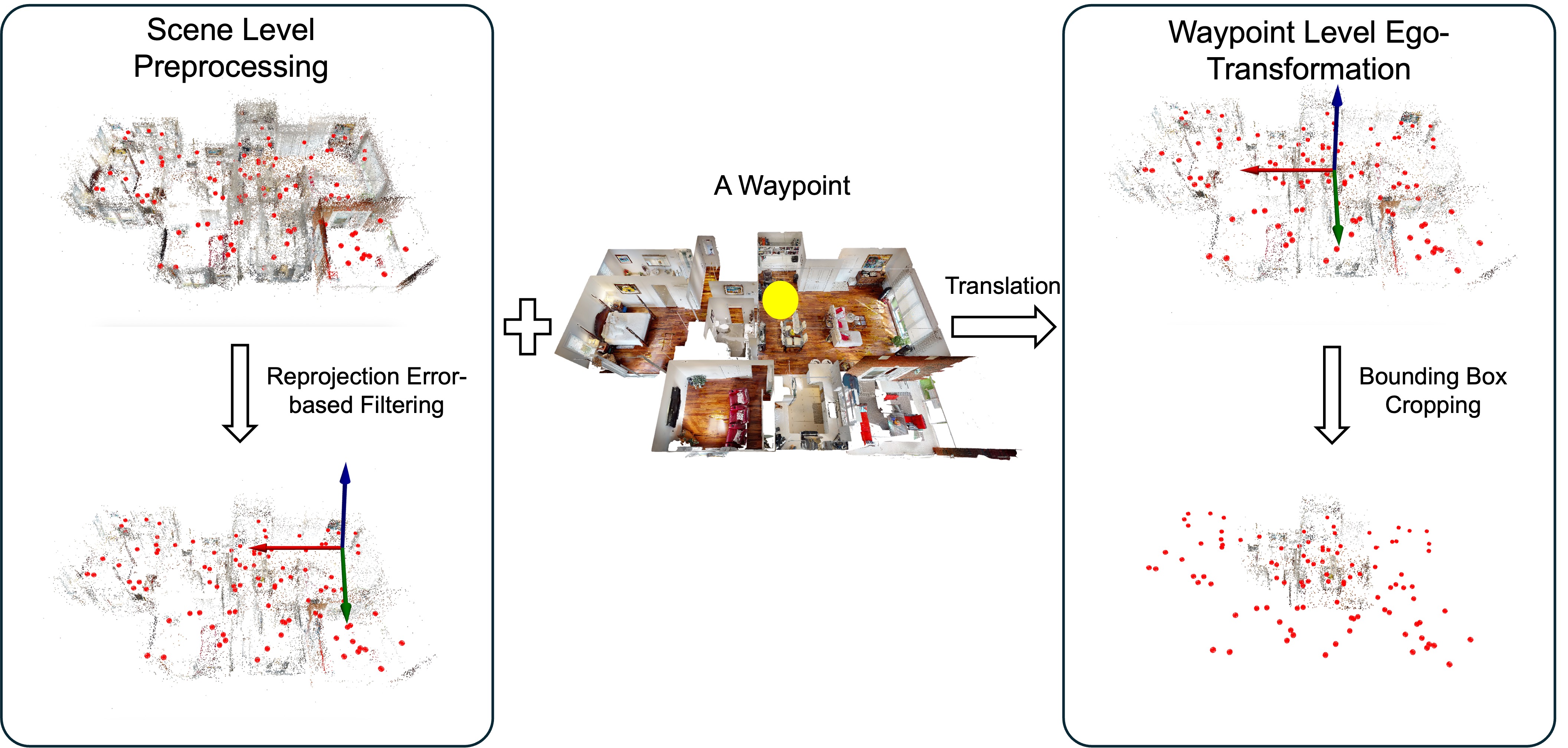} 
  \caption{\textbf{Data Preprocessing.} The pipeline removes highly uncertain 3D landmarks, applies an egocentric transformation to encode the waypoint position, and crops the point cloud within a bounding box centered at the waypoint. These steps enforce input consistency, focus the model on nearby information, and normalize the scale for training.}
  \label{fig:data_preprocessing_pipeline}
\end{figure}

The data preprocessing pipeline is demonstrated in \Cref{fig:data_preprocessing_pipeline}. This pipeline first removes highly uncertain points from the 3D landmarks and implicitly encodes the training waypoint coordinate into the data via egocentric transformation. Finally, it applies a bounding box cropping on the point cloud at the input waypoint to enforce the model to focus on nearby information, and also works as a normalization.

\section{Additional Experiments}
\subsection{Robustness to Sparse Reconstruction}
\label{sub:robustness_to_sparsification}

We evaluate the robustness of ActLoc under sparse reconstruction by progressively removing a percentage of camera poses and their associated 3D landmarks. 
We compare ActLoc against the baseline LWL on 10 HM3D test scenes. 
As shown in \Cref{tab:sparsification_results}, ActLoc maintains slightly higher localization success rates across most sparsification levels.

\begin{table*}[htbp]
  \makebox[\textwidth][c]{%
    \begin{tabular}{llrrrr}
      \toprule
      \textbf{Sparsification Level} & \textbf{Method}    & \textbf{0.1\,m/1\textdegree} & \textbf{0.25\,m/2\textdegree} & \textbf{0.5\,m/5\textdegree} & \textbf{5\,m/10\textdegree} \\
      \midrule
      0\%  & LWL               & 90.04 & 90.98 & 91.17 & 91.92 \\
           & ActLoc             & 92.11 & 92.48 & 92.86 & 93.23 \\
      \midrule
      10\% & LWL               & 89.47 & 90.23 & 90.41 & 91.54 \\
           & ActLoc             & 91.17 & 91.73 & 92.11 & 92.29 \\
      \midrule
      20\% & LWL               & 87.78 & 88.72 & 89.66 & 90.04 \\
           & ActLoc             & 90.79 & 91.54 & 92.29 & 92.67 \\
      \midrule
      30\% & LWL               & 88.53 & 89.66 & 90.23 & 90.79 \\
           & ActLoc             & 89.66 & 90.23 & 90.79 & 91.17 \\
      \midrule
      40\% & LWL               & 88.72 & 89.66 & 90.60 & 91.73 \\
           & ActLoc             & 90.23 & 90.98 & 91.17 & 91.54 \\
      \midrule
      50\% & LWL               & 89.85 & 90.41 & 90.98 & 91.35 \\
           & ActLoc             & 89.29 & 90.04 & 90.98 & 91.54 \\
      \midrule
      60\% & LWL               & 89.10 & 90.23 & 90.79 & 90.79 \\
           & ActLoc             & 88.35 & 89.66 & 90.04 & 90.41 \\
      \midrule
      70\% & LWL               & 87.78 & 88.72 & 89.29 & 89.85 \\
           & ActLoc             & 87.22 & 88.72 & 89.10 & 89.29 \\
      \midrule
      80\% & LWL               & 83.83 & 86.09 & 86.84 & 87.59 \\
           & ActLoc             & 86.84 & 87.59 & 88.16 & 88.91 \\
      \midrule
      90\% & LWL               & 83.08 & 84.02 & 84.77 & 85.53 \\
           & ActLoc             & 83.40 & 85.11 & 86.07 & 86.45 \\
      \bottomrule
    \end{tabular}%
  }  
  \caption{\textbf{Robustness under Sparse Reconstruction.} 
  Localization success rates (\%) under different sparsification levels, measured as the percentage of camera poses and associated points removed, on ten HM3D scenes. 
  ActLoc performs slightly better than LWL across most sparsification levels.}
  \label{tab:sparsification_results}
\end{table*}


\subsection{Extension to Multi-class Prediction}
Our model is designed with flexibility to adapt to different levels of prediction granularity. We implement a multi-class variant, ActLoc-Multi, where each viewpoint is assigned to one of four discrete quality levels instead of a binary label. All other training settings were kept the same as the binary model. As reported in \Cref{tab:actloc_multi_results}, this extension does not improve single-viewpoint selection performance. This aligns with our discussion in the main paper: simply increasing prediction granularity is not sufficient, and further system design, together with larger-scale datasets, may be needed to make such extensions effective.

\begin{table}[h]
\centering
\begin{tabular}{lcccc}
\toprule
 & 0.1m/1° & 0.25m/2° & 0.5m/5° & 5m/10° \\
\midrule
ActLoc & 92.11 & 92.48 & 92.86 & 93.23 \\
ActLoc-Multi & 88.35 & 89.47 & 89.66 & 90.41 \\
Upper Bound & 97.74 & 97.74 & 97.93 & 98.50 \\
\bottomrule
\end{tabular}
\vspace{2.0pt}
\caption{\textbf{Binary vs.\ Multi-class ActLoc.} 
Comparison of binary and multi-class models on the HM3D test set, reported as localization success rates (\%) at different thresholds in the single-viewpoint selection task. 
The binary model consistently outperforms the multi-class variant, suggesting that finer prediction granularity does not necessarily improve performance.}
\label{tab:actloc_multi_results}
\end{table}

\subsection{Experiments on Focal Axis Rotation}
\label{subsec:experiments_on_focal_axis_rotation}

We evaluate the effect of in-plane (roll) rotations on visual localization using 739 viewpoints from 12 HM3D scenes. 
At each viewpoint, the virtual camera is rotated around its principal axis in $90^\circ$ increments, and localization is performed at each angle. 
As summarized in \Cref{tab:inplane_rotation}, large roll angles (e.g., $\pm90^\circ$ and $180^\circ$) result in extremely poor success rates, while the best performance is consistently observed at $0^\circ$ (no in-plane rotation). 
These results confirm our decision to exclude in-plane rotations from the LocMap representation.

\begin{table}[t]
\centering
\setlength{\tabcolsep}{6pt}
\begin{tabular}{lcccc}
\toprule
Roll angle (deg) & 0.1m/1° & 0.25m/2° & 0.5m/5° & 5m/10° \\
\midrule
-90 & 0.14 & 0.14 & 0.14 & 0.41 \\
\textbf{0}   & \textbf{84.17} & \textbf{86.20} & \textbf{87.55} & \textbf{87.96} \\
90  & 0.27 & 0.27 & 0.27 & 0.27 \\
180 & 0.41 & 0.95 & 1.08 & 1.35 \\
\bottomrule
\end{tabular}
\caption{\textbf{Effect of In-plane Rotations.} Localization success rates (\%) under different roll angles, evaluated on 739 viewpoints from 12 HM3D scenes. Large roll angles (e.g., $\pm90^\circ$ and $180^\circ$) lead to extremely poor localization, confirming our decision to exclude in-plane rotations from LocMap.}
\label{tab:inplane_rotation}
\end{table}

\subsection{Scalability to Higher-Resolution LocMaps}
\label{subsec:experiments_on_large_grid_resolutions}

We evaluate inference efficiency under different LocMap resolutions by comparing ActLoc with LWL. 
Both methods are tested at 6$\times$18, 12$\times$36, and 24$\times$72 output grids. 
As shown in \Cref{tab:inference_time_comparison}, ActLoc requires about 108\,ms to process a single waypoint, with less than 1\,ms variation across runs and resolutions. 
In contrast, LWL requires on average 8.3\,s, 27.5\,s, and 104.4\,s at the three resolutions, respectively. 
This corresponds to a speed-up of two to three orders of magnitude, demonstrating that ActLoc scales efficiently to higher-resolution LocMaps.

\begin{table}[t]
\centering
\begin{tabular}{lccc}
\toprule
Method & 6$\times$18 & 12$\times$36 & 24$\times$72 \\
\midrule
LWL    & 8256 $\pm$ 56 & 27450 $\pm$ 92 & 104354 $\pm$ 545 \\
\textbf{ActLoc} & \textbf{108.4 $\pm$ 0.7} & \textbf{108.8 $\pm$ 0.2} & \textbf{109.0 $\pm$ 0.9} \\
\bottomrule
\end{tabular}
\caption{\textbf{Scalability to Higher-Resolution LocMaps.} 
Average processing time per waypoint with standard deviation (ms). 
ActLoc maintains a nearly constant runtime across resolutions ($\approx$108\,ms), while LWL increases from $\sim$8\,s to over 100\,s, resulting in a $10^2$–$10^3$ speedup.}
\label{tab:inference_time_comparison}
\end{table}

\subsection{Scalability to Large Scenes}
\label{subsec:experiments_on_large_scenes}

We evaluate the scalability of ActLoc to large scenes by comparing its average processing time per waypoint against LWL. 
Scene size is measured by the number of SfM landmarks after error filtering. 
We divide the scenes into three ranges: $[0, 50k)$, $[50k, 100k)$, and $[100k, 150k)$. 
For each range, three scenes are randomly sampled, and each experiment is repeated 3 times. 
As shown in \Cref{tab:scene_size}, the runtime of LWL increases rapidly with scene size, whereas ActLoc remains efficient, demonstrating strong scalability to larger environments.

\begin{table}[h]
\centering
\begin{tabular}{lccc}
\toprule
Method & $<50k$ & $50k$--$100k$ & $100k$--$150k$ \\
\midrule
LWL    & 7249 & 13803 & 25650 \\
\textbf{ActLoc} &   \textbf{93} &   \textbf{125} &   \textbf{191} \\
\bottomrule
\end{tabular}
\caption{\textbf{Scalability to Large Scenes.} 
Average processing time per waypoint (ms, rounded to the nearest integer) under different scene sizes, measured by the number of SfM landmarks after error filtering. 
ActLoc remains efficient, while LWL runtime grows rapidly with scene size.}
\label{tab:scene_size}
\end{table}

\subsection{Cross-domain Generalization with Alternative Features}

We evaluate the cross-domain generalization of ActLoc by replacing the default SuperPoint features with SIFT. 
Experiments are conducted on five ScanNet scenes, comparing ActLoc and the LWL baseline. 
As shown in \Cref{tab:sift_scannet_results}, ActLoc does not outperform LWL when using SIFT features. 
This result highlights a trade-off in our design: although processing raw inputs enables high efficiency, it can reduce robustness to alternative feature types.

\begin{table}[h]
\centering
\begin{tabular}{lcccc}
\toprule
 & 0.1m/1° & 0.25m/2° & 0.5m/5° & 5m/10° \\
\midrule
LWL \cite{di2024learning} & 36.36 & 36.36 & 36.36 & 36.36 \\
ActLoc & 22.22 & 22.22 & 22.22 & 22.22 \\
Upper Bound & 56.57 & 57.58 & 57.58 & 60.61 \\
\bottomrule
\end{tabular}
\vspace{2.0pt}
\caption{\textbf{Cross-domain Generalization with SIFT.} 
Performance of SIFT-based visual localization across five ScanNet scenes, reported as localization success rates (\%) at different thresholds. 
Unlike results with SuperPoint, ActLoc does not outperform LWL in this setting, highlighting reduced robustness to alternative feature types.}
\label{tab:sift_scannet_results}
\end{table}

\subsection{Effect of LocMap Resolution}
\label{subsec:effect_of_LocMap_resolution}

We adopt a $6\times18$ resolution for the LocMap as a trade-off between efficiency and accuracy. 
Note that ActLoc itself is trained with binary labels obtained by thresholding localization errors, 
while here we visualize the underlying continuous error maps. 
This provides a more detailed view of how localization difficulty varies with viewpoint resolution, 
from which the binary supervision is derived.  

\begin{figure}[t]
  \centering
  \includegraphics[width=\linewidth]{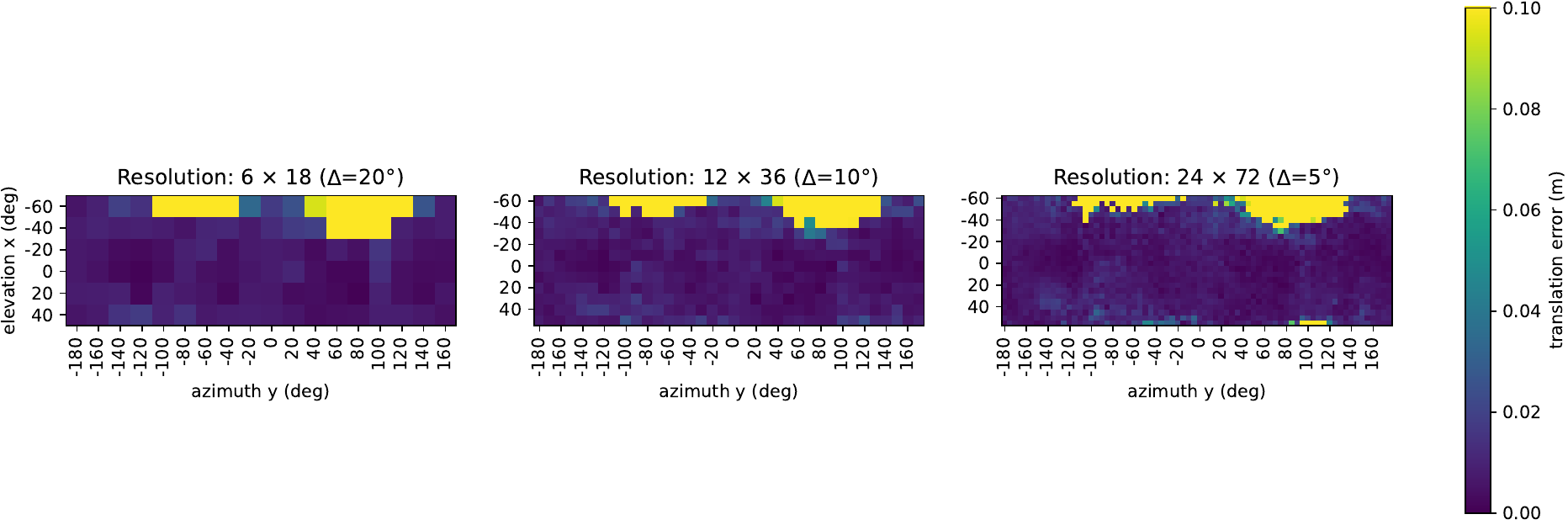}
  \caption{\textbf{Consistency Across Resolutions.} 
  Localization error maps at resolutions of $6\times18$, $12\times36$, and $24\times72$ (left to right). 
  Overall error patterns remain consistent across resolutions, showing that coarse maps already capture the main spatial structure. Translation errors are capped at $0.1$\,m for visualization.}
  \label{fig:resolution_comparison}
\end{figure}

We first compare ground-truth error maps at $6\times18$, $12\times36$, and $24\times72$ resolutions. 
As shown in \Cref{fig:resolution_comparison}, the overall error patterns remain consistent across resolutions, 
indicating that coarse grids already capture the dominant spatial structure. 
For brevity, we only show translation error; rotation error exhibits similar trends.  

\begin{figure}[t]
  \centering
  \includegraphics[width=\linewidth]{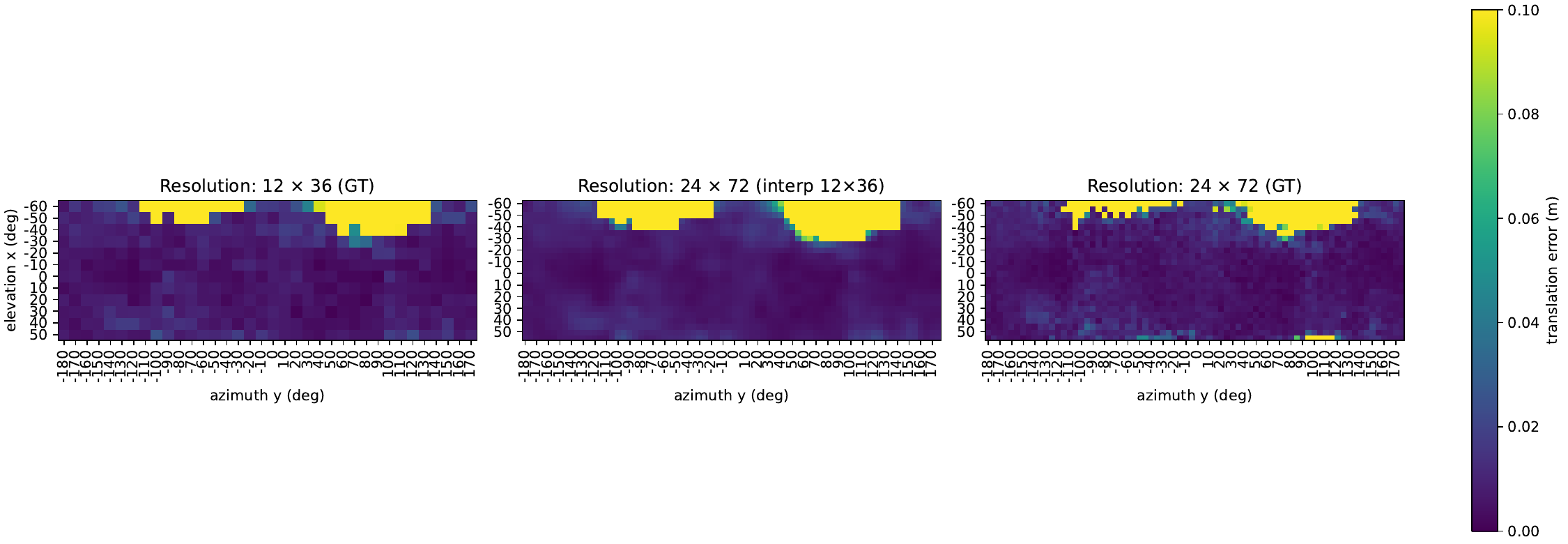}
  \caption{\textbf{Interpolation to Higher Resolutions.} 
  Error maps from ground-truth $12\times36$, interpolated $24\times72$, and ground-truth $24\times72$ (left to right). 
  Interpolated maps closely resemble the true high-resolution results, supporting upsampling as a viable strategy.}
  \label{fig:interpolation_comparison}
\end{figure}

We further evaluate whether higher-resolution maps can be approximated via interpolation. 
\Cref{fig:interpolation_comparison} compares the ground-truth $12\times36$ and $24\times72$ maps with an interpolated $24\times72$ map derived from the $12\times36$ version. 
The interpolated result closely matches the ground-truth $24\times72$ map, 
suggesting that ActLoc predictions at low resolution can be upsampled to higher resolutions with minimal loss.

\section{Qualitative Results}
\subsection{LocMap-based Viewpoint Prediction Visualization}

\begin{figure}[htbp]
  \centering
  \includegraphics[width=0.8\linewidth]{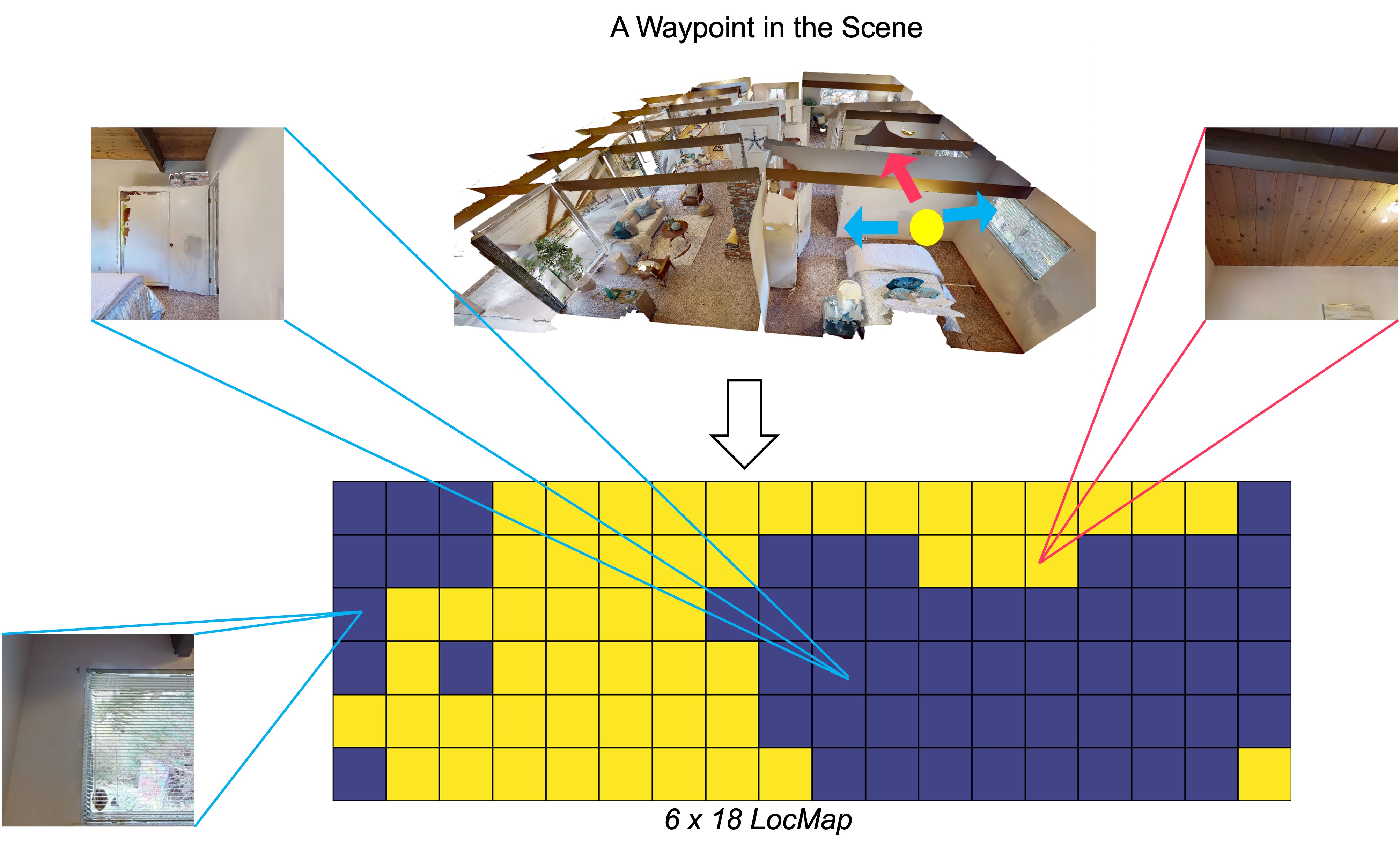} 
  \includegraphics[width=0.8\linewidth]{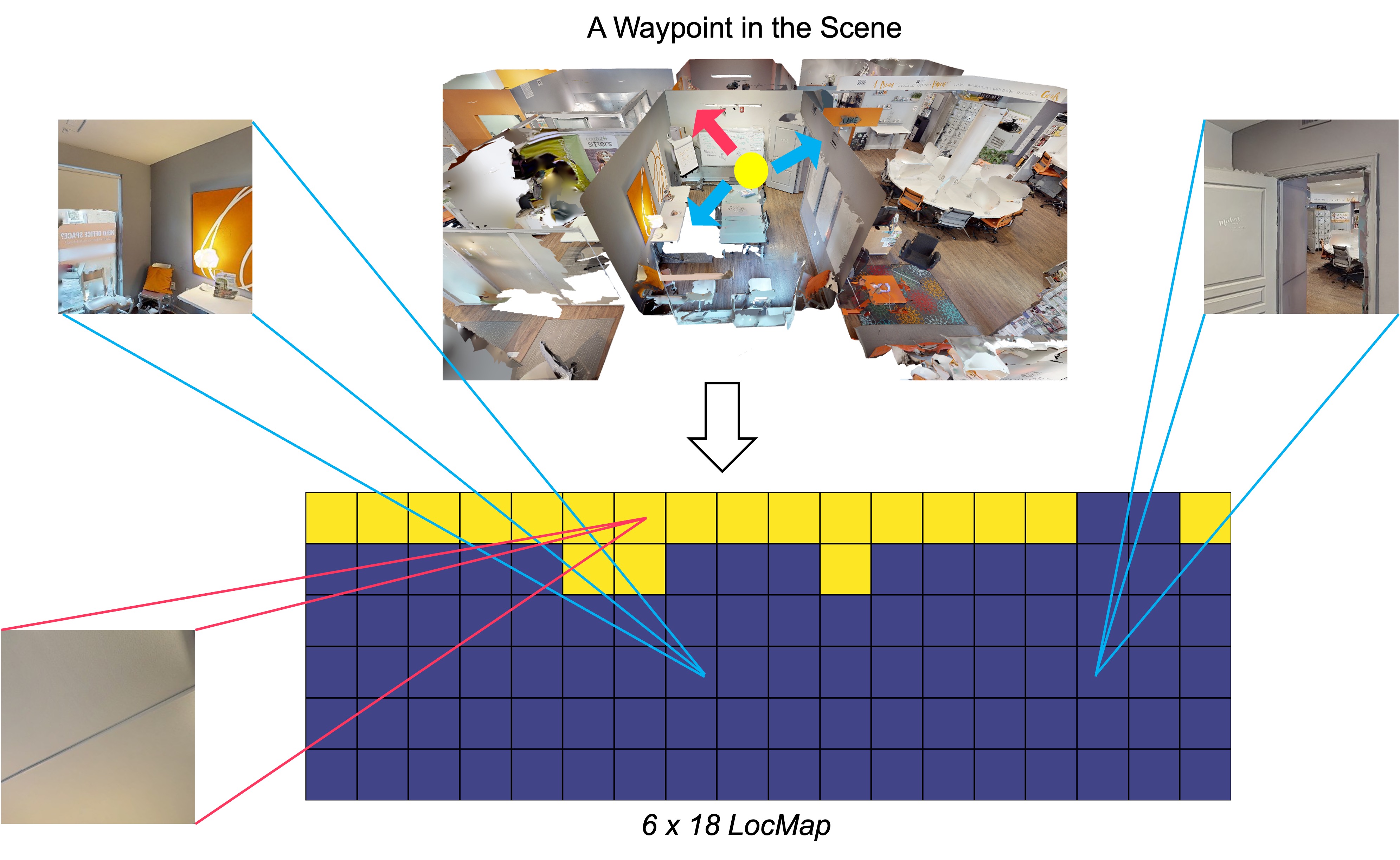} 
  \includegraphics[width=0.8\linewidth]{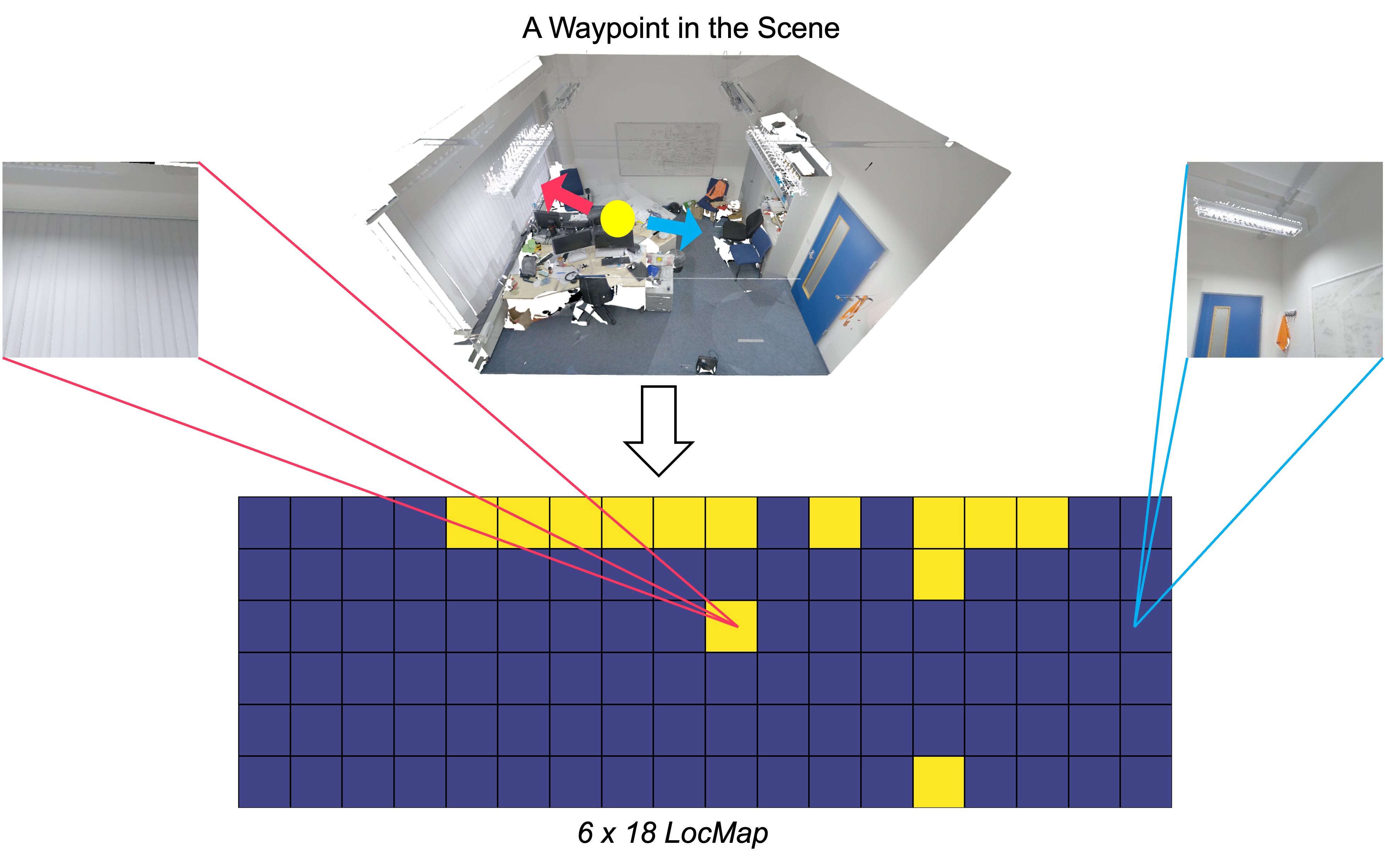} 
  \caption{\textbf{Visualization of Viewpoint Prediction.} 
Examples from HM3D and ScanNet. 
Each figure shows a scene mesh with a waypoint and candidate viewing directions (arrows), the corresponding $6 \times 18$ LocMap predictions (deep indigo = easy to localize, bright yellow-green = hard to localize), and sample images for selected viewpoints.}
  \label{fig:viewpoint_selection_vis}
\end{figure}

We visualize examples of viewpoint prediction from our LocMap model in \Cref{fig:viewpoint_selection_vis}. 
Each example shows a scene mesh with a waypoint at the center, where arrows indicate candidate viewing directions. 
Below, the predicted LocMap is shown as a $6 \times 18$ grid, where deep indigo cells denote viewpoints predicted as easy to localize and bright yellow-green cells as hard to localize. 
For the candidate viewpoints highlighted in the scene, their corresponding cells in the LocMap are connected to the actual captured images, providing an intuitive link between the prediction and scene appearance.

\subsection{Path Planning Visualization}

\begin{figure}[htbp]
  \centering
  \includegraphics[width=0.99\linewidth]{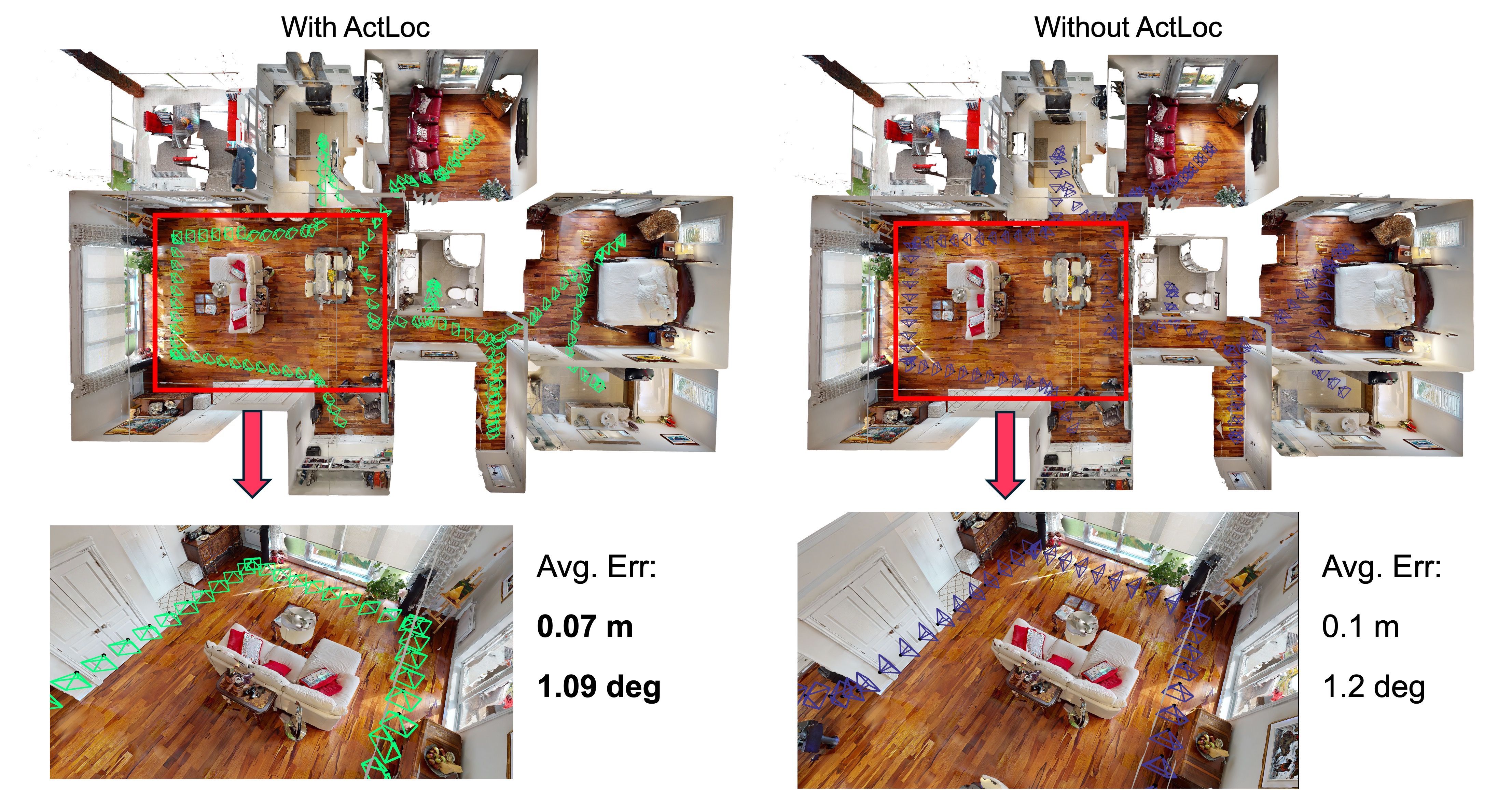}
  \includegraphics[width=0.99\linewidth]{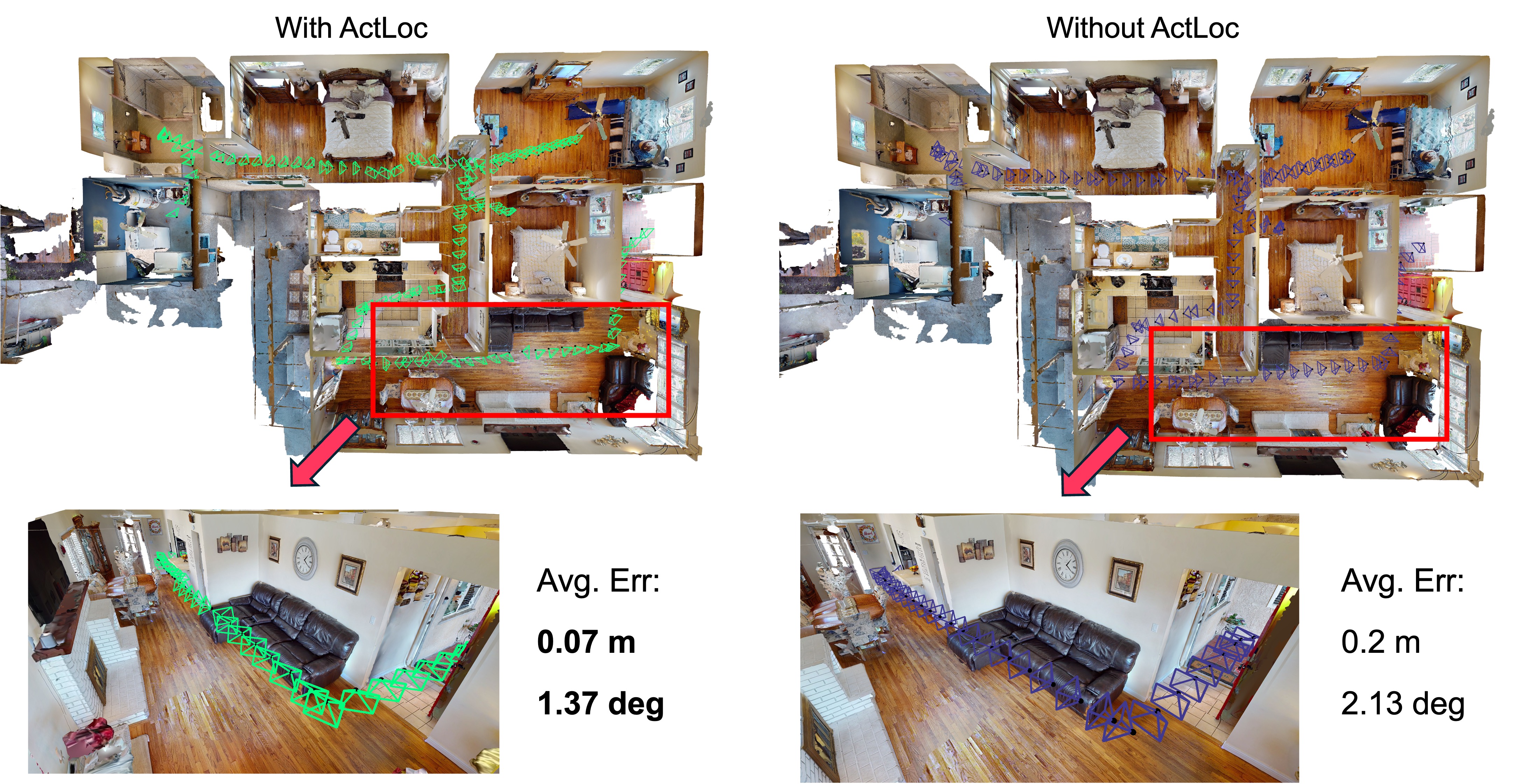}
  \caption{\textbf{Path Planning Visualization.} 
Comparison between ActLoc (left) and a forward-facing baseline (right) in two HM3D test scenes (top and bottom). 
Each viewpoint along the path is shown as a camera frustum in a top-down view. 
Regions with clear differences are highlighted with red boxes and further shown in zoom-in views. 
For each scene, the average translation and rotation errors are also reported, where ActLoc consistently achieves lower errors than the baseline.}

  \label{fig:path_planning_vis1}
\end{figure}

We provide additional visualizations of path planning results in two HM3D test scenes, shown in \Cref{fig:path_planning_vis1}. 
In each scene, we compare ActLoc against a forward-facing baseline, where planned viewpoints are shown as camera frusta in a top-down view. 
Regions with large differences are highlighted and further shown in zoom-in views, and for each scene the average translation and rotation errors are also reported. 
These examples illustrate how ActLoc selects viewpoints that improve localization accuracy while maintaining feasible paths.

\subsection{Global Score Heatmap from ActLoc-Multi}
\label{sub:global_score_heatmap}

\begin{figure}[htbp]
  \centering
  \includegraphics[width=1.0\textwidth]{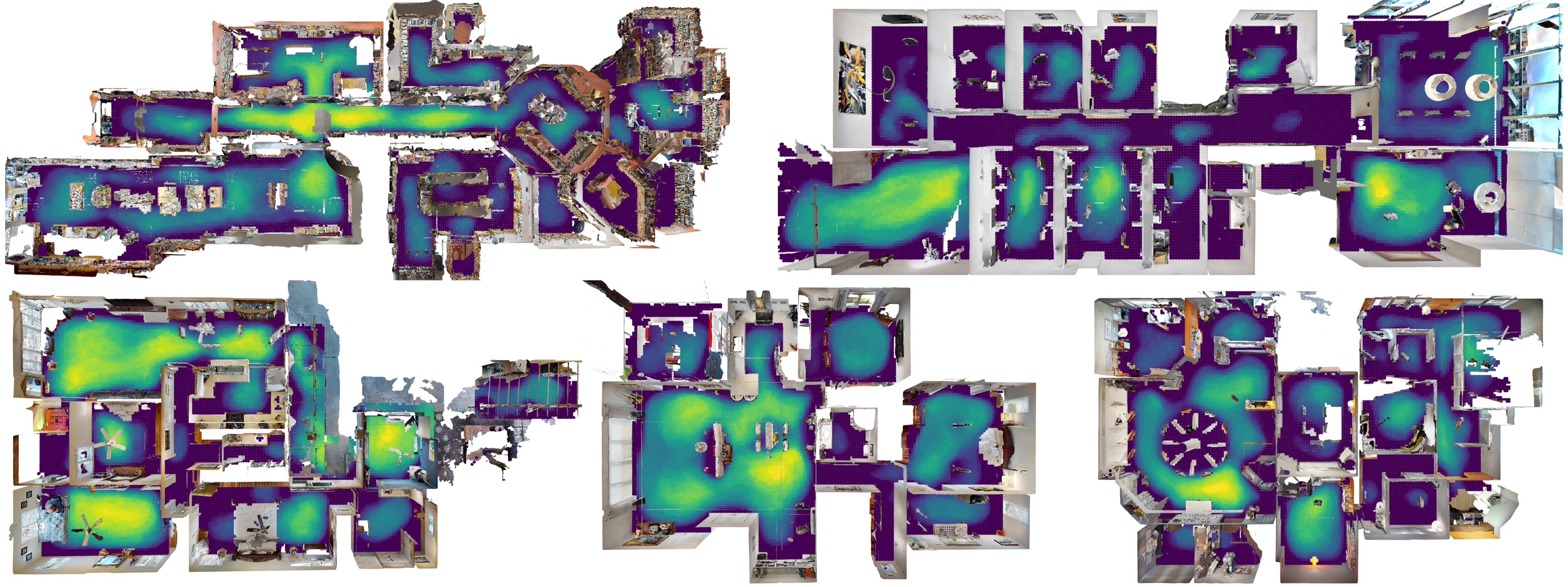}
  \caption{\textbf{Global Localization Score Maps.} 
Heatmap visualizations generated by ActLoc-Multi, showing averaged scores across dense waypoints in different indoor scenes. 
Feature-rich regions, such as areas with more furniture, exhibit higher scores, highlighting the utility of the multi-class model for capturing spatial variations in localization difficulty.}
  \label{fig:heatmap}
\end{figure}

To illustrate the utility of the multi-class variant, we generate global localization score maps using ActLoc-Multi. 
We sample dense waypoints at a height of 0.5\,m, run the model at each waypoint, and average the top five grid scores to obtain a scalar value. 
The resulting values are visualized as heatmaps in \Cref{fig:heatmap}. 
These maps reveal clear spatial variations: feature-rich regions (e.g., areas with more furniture) tend to have higher scores, indicating that ActLoc-Multi can still provide useful global cues for planning.

